\documentclass[conference]{IEEEtran}
\IEEEoverridecommandlockouts

\usepackage{cite}
\usepackage{amsmath,amssymb,amsfonts}
\usepackage{algorithmic}
\usepackage{graphicx}
\usepackage{textcomp}
\usepackage{xcolor}
\usepackage{multirow}
\usepackage[inkscapearea=page]{svg}
\usepackage{adjustbox}

\def\BibTeX{{\rm B\kern-.05em{\sc i\kern-.025em b}\kern-.08em
    T\kern-.1667em\lower.7ex\hbox{E}\kern-.125emX}}
    
\usepackage[bookmarks=false]{hyperref}
    \hypersetup{colorlinks,
      linkcolor=blue,
      citecolor=green,
      urlcolor=blue}

\usepackage{graphics} 
\usepackage{dirtree}
\usepackage[capitalize]{cleveref}
\graphicspath{ {./images/} }
\usepackage{subcaption}
\usepackage{multicol}

\makeatletter 
\newcommand{\linebreakand}{%
  \end{@IEEEauthorhalign}
  \hfill\mbox{}\par
  \mbox{}\hfill\begin{@IEEEauthorhalign}
}
\makeatother 

\begin{document}

\title{Safety-Driven Deep Reinforcement Learning  Framework  for Cobots: A Sim2Real Approach}

\author{\IEEEauthorblockN{1\textsuperscript{st} Ammar N. Abbas}
\IEEEauthorblockA{\textit{Data Science} \\
\textit{Software Competence Center}\\
Hagenberg, Austria \\
ammar.abbas@scch.at}
\and
\IEEEauthorblockN{2\textsuperscript{nd} Shakra Mehak}
\IEEEauthorblockA{\textit{Industrial Automation} \\
\textit{Pilz}\\
Cork, Ireland \\
s.mehak@pilz.ie}
\and
\IEEEauthorblockN{3\textsuperscript{rd} Georgios C. Chasparis}
\IEEEauthorblockA{\textit{Data Science} \\
\textit{Software Competence Center}\\
Hagenberg, Austria \\
georgios.chasparis@scch.at}
\and

\IEEEauthorblockN{4\textsuperscript{th} John D. Kelleher}
\IEEEauthorblockA{\textit{Computer Science} \\
\textit{Trinity College Dublin}\\
Dublin, Ireland \\
john.kelleher@tcd.ie}
\and

\linebreakand 

\IEEEauthorblockN{5\textsuperscript{th} Michael Guilfoyle}
\IEEEauthorblockA{\textit{Industrial Automation} \\
\textit{Pilz}\\
Cork, Ireland \\
m.guilfoyle@pilz.ie}
\and
\IEEEauthorblockN{6\textsuperscript{th} Maria Chiara Leva}
\IEEEauthorblockA{\textit{Food Science and Environmental Health} \\
\textit{Technological University Dublin}\\
Dublin, Ireland \\
mariachiara.leva@tudublin.ie}
\and
\IEEEauthorblockN{7\textsuperscript{th} Aswin K Ramasubramanian}
\IEEEauthorblockA{\textit{Robotics and Automation} \\
\textit{Irish Manufacturing Research}\\
Mullingar, Ireland \\
aswin.ramasubramanian@imr.ie}
}\maketitle
\begin{abstract}

This study presents a novel methodology incorporating safety constraints into a robotic simulation during the training of deep reinforcement learning (DRL). The framework integrates specific parts of the safety requirements, such as velocity constraints, as specified by ISO 10218, directly within the DRL model that becomes a part of the robot's learning algorithm. The study then evaluated the efficiency of these safety constraints by subjecting the DRL model to various scenarios, including grasping tasks with and without obstacle avoidance. The validation process involved comprehensive simulation-based testing of the DRL model’s responses to potential hazards and its compliance. Also, the performance of the system is carried out by the functional safety standards IEC 61508 to determine the safety integrity level. The study indicated a significant improvement in the safety performance of the robotic system. The proposed DRL model anticipates and mitigates hazards while maintaining operational efficiency. This study was validated in a testbed with a collaborative robotic arm with safety sensors and assessed with metrics such as the average number of safety violations, obstacle avoidance, and the number of successful grasps. 
The proposed approach outperforms the conventional method by a 16.5\% average success rate on the tested scenarios in the simulations and 2.5\% in the testbed without safety violations. The project repository is available at \href{https://github.com/ammar-n-abbas/sim2real-ur-gym-gazebo}{https://github.com/ammar-n-abbas/sim2real-ur-gym-gazebo}.
\end{abstract}


\section{Introduction}
Deep Reinforcement Learning (DRL) offers potential within Human-Robot Collaboration (HRC), yet its adoption within real-world industrial robotics is constrained by safety concerns due to its interaction with operators \cite{li2024unleashing}. However, these safety concerns are overcome by the integration of safety-rated control systems such as PLC, relays, e-stops, and safety scanners which adhere to industrial safety standards. Thereby with appropriate training strategies, a DRL algorithm can be tuned to provide dynamic and adaptable capabilities that can offer effective solutions to address diverse challenges, particularly in unobserved and unexpected situations that demand extreme caution while incorporating safety protocols \cite{garcia2015comprehensive, thumm2022provably}. For instance, the need to efficiently transfer a learned policy from a simulation environment to the real world, keeping the behavior of the system robust across different conditions and contexts is safety critical \cite{qian2014manipulation}. This implies the need for integrating safety standards into the learning framework that ensures the DRL algorithm complies with the regulatory and safety requirements. While also ensuring the reliability of DRL-driven robotic systems across various sectors. Moving forward, the traditional DRL-based approaches can be modified with the integration of safety protocols into the learning algorithm to change from fixed rule-driven safety to a flexible learning-based method. 

\begin{figure}[t]
  \centering
  \includegraphics[width=\linewidth]{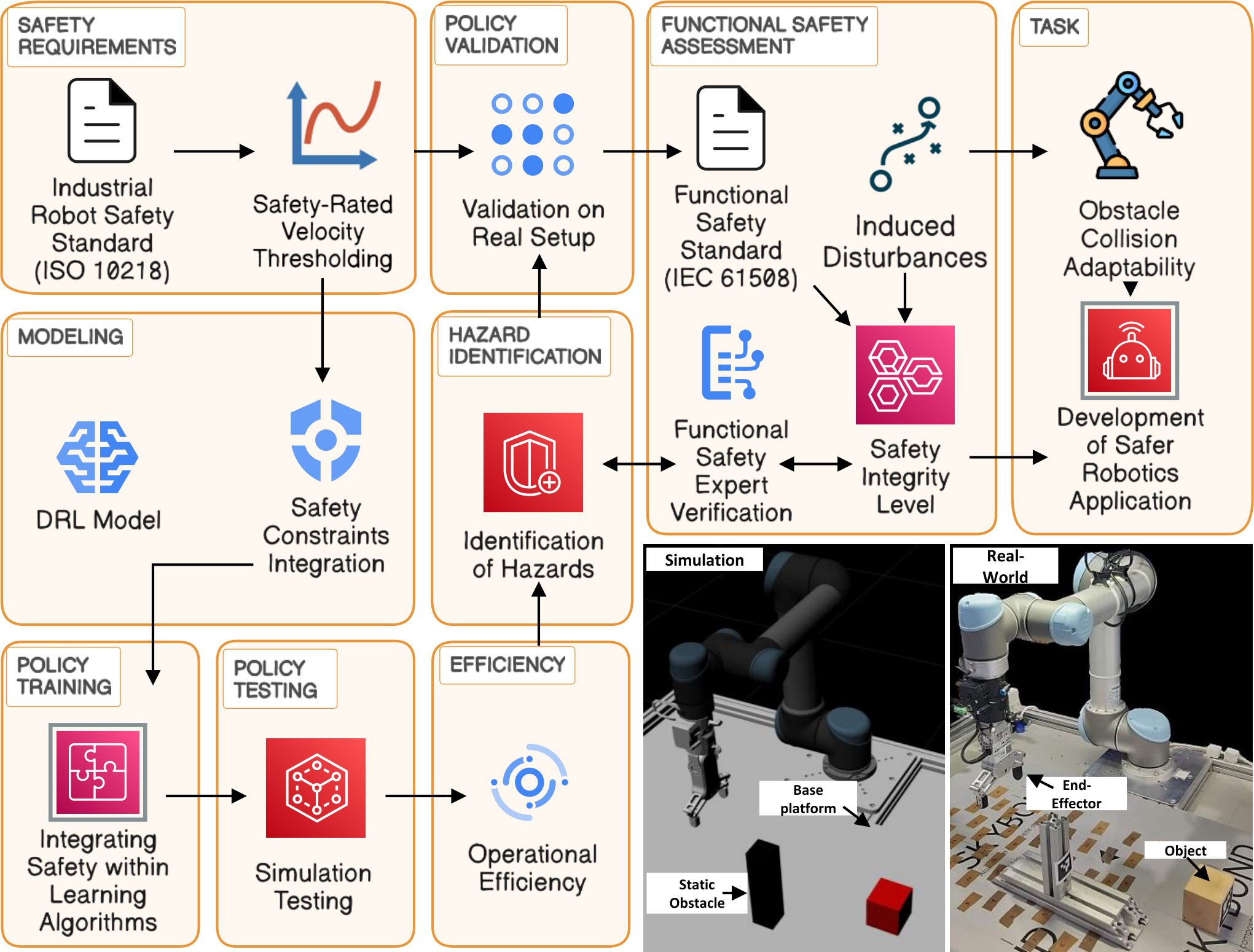}
  \caption{Safety-Driven Deep Reinforcement Learning.}
  \label{fig:framework}
  \vspace{-0.5cm}
\end{figure}


Central to this effort are the guidelines outlined in ISO 10218 \cite{DINENISO10218-1_2012}, which cover the safety regulations for industrial robots, and ISO/IEC 61508 \cite{DINEN61508-1_2016}, which ensures the functional safety of electrical/electronic/programmable electronic safety-related systems are considered to be key in robotic system evaluations \cite{tong2017systematic, chen2017verification}. These regulations offer a solid groundwork for guaranteeing that the robotic systems are designed, implemented, and managed with a primary focus on enhancing safety. DRL-based algorithms developed in collaborative cells necessitate a safe robotic operation, therefore, functional safety evaluation should be considered a part of the process. These DRL algorithms are capable of encountering an undesirable scenario and may carry out an operation that may compromise the safety of the system. Moreover, integrating safety into DRL involves developing a reward function that balances operational efficiency and complies with safety regulations. Conventional approaches, which frequently translate optimization objectives into rewards, involve modifications to assist robots in navigating intricate and potentially inconsistent goals. Sparse reward functions are frequently employed in DRL for robotics to promote the unbiased acquisition of optimal methods \cite{krasowski2023provably}. However, these functions often need to consider the safety measures that are essential for the setup in real-world applications.

This study introduces a novel framework named \textit{``Safety-Driven DRL (SD-DRL)"} as shown in \cref{fig:framework}, which builds on traditional DRL with the ISO 10218 and IEC 61508 functional safety assessment and compares its performance against traditional DRL model. The proposed framework, SD-DRL is designed to ensure the safe operation of the collaborative robotic cell with an ISO-compliant DRL algorithm. Following the development and validation of the DRL framework in a simulation, the program is transferred to a real robotic cell (via the Sim2Real approach). This approach is then evaluated and assigned safety integrity levels (SILs). This step ensures that each function within the robotic system is classified according to its risk and the necessary level of safety reliability. Thus proposed DRL represents a step further toward the deployment of safer, more reliable robotic systems. The structure of the paper consists of \cref{sec:sota}: State-of-the-art, followed by \cref{sec:meth}: Methodology, \cref{sec:doe}: Design of experiments, \cref{sec:results}: Experimental results and discussion, \cref{sec:fsa}: Functional safety assessment, and \cref{sec:conc}: Conclusions and future work.

\section{State-of-the-Art}
\label{sec:sota}
Industrial robotics standards mainly ISO 10218 \cite{DINENISO10218-1_2012} provide an emphasis on robotic safety by defining the system requirements for robots and robotic systems with a key focus on reducing the possibility of causing injury or damage to humans and the environment. 

The literature presents an increase in the integration of safety standards in the design and management of robotic systems \cite{lakshminarayanan2024robots}, focusing on ensuring compliance through various means, from design principles to operational protocols. The latest developments in safe-DRL components related to task planners \cite{krasowski2023provably} have enhanced agents' behavior in robotic scenarios for the unstructured operational environment to seamlessly interact. This enables safety requirements to be incorporated into the robot's learning objectives for further enhancing the autonomy level\cite{marchesini2022exploring,hsu2022improving}.


Integration of safety constraints into DRL has always been a challenging task in terms of implementing safety requirements into the learning algorithm \cite{brunke2022safe,shao2021reachability}. There are two primary categories of DRL algorithms in which either the optimization goals or the learning processes \cite{fan2024learn,el2020towards} are adjusted. This involves developing agents that can avoid dangerous states, ensuring safety even in less-than-ideal conditions such as obstacle avoidance \cite{sangiovanni2020self}. Several studies suggest that the safe agents in DRL depend on the capacity to make decisions, infer, and adjust in alignment with human preferences \cite{kim2022automating,lange2012autonomous,gupta2021reset}. The authors in \cite{thomas2021safe} introduce a criterion for agent safety that allows agents to move from any state to another, assuring error recovery. The approach emphasizes the significance of creating agents that can perform reversible behaviors to improve their safety in unpredictable circumstances. Additionally, \cite{hunt2021verifiably} delves into the concept of secure exploration in robot reinforcement learning, and \cite{liu2023safe} reformulates the exploration procedure within the tangent space of the constraint manifold. This modification alters the agent's action space to comply with safety constraints on a local scale continuously. In the study conducted by \cite{zhang2023safety}, the integration of control barrier functions into control policies is introduced to enforce safety constraints, guaranteeing that robotic systems function within secure boundaries.

The importance of safety shielding and reward function shaping in DRL for ensuring safety while maintaining task performance is emphasized in recent research \cite{alshiekh2018safe,jansen2020safe}. The authors in \cite{ray2019benchmarking,garcia2015comprehensive,yang2023safety} underscore the significance of reward engineering in guiding robots toward desired behaviors while upholding safety standards. For instance, \cite{plappert2018multi} presented a multi-goal reward function using hindsight experience replay, which enables learning from unsuccessful episodes by redefining goals. They use dense and sparse rewards to develop their reward function to evaluate agent performance, and the reward function focuses solely on goal achievement without penalizing unsafe actions. Moreover, the Safety-Gymnasium benchmark \cite{ji2023safety} aims to provide standardized evaluation environments for SafeRL, focusing on safety-critical tasks. Another study \cite{saliba2018training} presents a training simulator integrating Gazebo \cite{qian2014manipulation} and Robot Operating System (ROS) \cite{koubaa2017robot} to control robot models, highlighting its adaptability and usability across various scenarios. Several studies have compared the performance of Open Dynamics Engine (ODE) with other physics engines such as Vortex, Bullet, MoJoco, and PhyX, in Gazebo simulator and CoppeliaSim \cite{yoon2023comparative,connolly2021realistic}. Based on the availability in Gazebo, ODE was used as the physics engine in this study. 

This paper presents several key contributions including a Sim2Real validation (similar to \cite{ramasubramanian2022automatic}). Firstly, building on top of \cite{1802.09464,ur_cambel}, a simulation environment and benchmark for industrial robotics and ROS-based platforms to validate safety constraints and facilitate simulation-to-reality transfer is proposed in this study using Gazebo and Gymnasium-Robotics. Secondly, it describes the development of a safety-driven DRL reward function, which incorporates the safety-rated reduced speed consideration of $250mm/s$ at collision taken from ISO 10218. Thirdly, it evaluates the performance of the proposed safety-driven DRL model against a traditional DRL model \cite{katyal2017leveraging}, across both simulated and real-world scenarios, showcasing its reliability and efficacy. Finally, the paper outlines a functional safety assessment conducted following the ISO/IEC 61508 standard, with validation by an industrial functional safety expert at Pilz. This assessment assigns Safety Integrity Levels (SILs) to the DRL application, with simulation testing and real-world validation conducted to verify its functional safety. 

\section{Methodology}
\label{sec:meth}
This section discusses the methods and the development of SD-DRL and its evaluation. Firstly, the environment that involves the software platform is used to bridge the communication between the algorithm and the simulated and real-world robot setup. Secondly, the task space used for the evaluation. Further, the characteristics of the DRL model are discussed along with its state, action, and reward. Thirdly, the evaluation strategies are presented. Finally, the safety standards validation strategy. 

\subsection{Environment Framework}
The environment (built on top of \cite{1802.09464,ur_cambel}) in this case study includes the simulated and real-world settings where the robotic tasks are validated. The simulation workspace is available at the project \href{https://github.com/ammar-n-abbas/sim2real-ur-gym-gazebo.git}{repository}\footnote{\href{https://github.com/ammar-n-abbas/sim2real-ur-gym-gazebo.git}{https://github.com/ammar-n-abbas/sim2real-ur-gym-gazebo.git}} \cite{ammar_n_abbas_2024_10569005}. 

This study employs the "Universal Robot Grasp Task Space \texttt{(UR5GraspEnv-v0)}" as the robotic workspace, terminating the episodes upon collisions with the environment or successful grasp. It utilizes the ROS \cite{koubaa2017robot} as a middleware framework for communication and control while Gazebo as the simulator \cite{qian2014manipulation} for testing and development of the algorithm before deployment onto the real UR5 robot testbed \cite{jiang2021vision, wang2020learning}. Evaluation and benchmarking of reinforcement learning algorithms for robotic tasks proposed in the study are conducted using the Gymnasium-Robotics environment \cite{gymnasium_robotics2023github}. 

\begin{figure}[tbp]
  \centering
  \includegraphics[width=\linewidth]{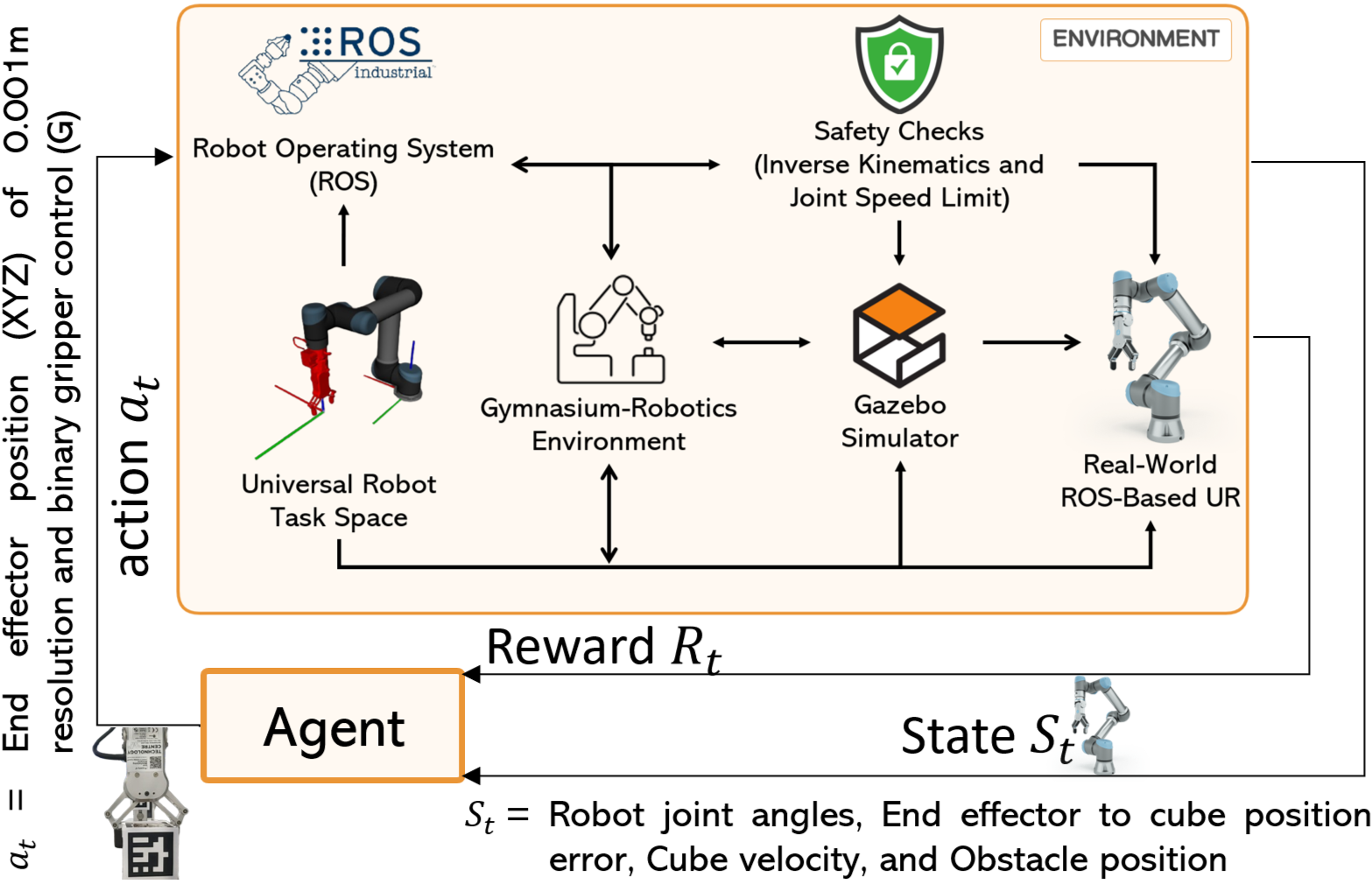}
  \caption{Sim2Real environment framework.}
  \label{fig:sim2real}
  \vspace{-0.5cm}
\end{figure}

Simulation-to-Reality (Sim2Real) approach aims to train, test, and transfer models from simulated environments to real-world applications, thereby saving significant time \cite{ramasubramanian2022automatic}. Such approaches can be scaled to address the challenges in deploying learned DRL policies onto physical robots after being validated in simulation. Similarly, this study focuses on enabling zero-shot transfer to deploy trained policies from simulation to physical robots without additional fine-tuning. The proposed architecture is illustrated in \cref{fig:sim2real}.

\subsection{Deep Reinforcement Learning}
Due to its adaptive learning capabilities, DRL has the ability to solve complex decision-making issues.

\subsubsection{Algorithm and Hyperparameters}
The study uses Truncated Quantile Critics (TQC) \cite{kuznetsov2020controlling} as our preferred DRL algorithm, which improves DRL's limitations in addressing overestimation bias. 

\subsubsection{State}
The \texttt{state} ($S_{t}$) in the \texttt{UR5GraspEnv} environment is shown in \cref{fig:sim2real}.


\subsubsection{Action}
The 4-dimensional \texttt{action} ($a_{t}$) vector in the \texttt{UR5GraspEnv} environment is shown in \cref{fig:sim2real}. 

\subsubsection{Reward}
The traditional DRL's \texttt{reward} function (\cref{eq:rew}) uses a standard approach in such similar robotic task environments \cite{1802.09464}. The study has introduced a novel modified reward function as proposed in the framework of SD-SDRL, incorporating safety which aims to encourage the agent's behaviors to result in successful goal achievement while maintaining safe operation, expressed mathematically in \cref{eq:mod_rew}.
\begin{equation}
\label{eq:rew}
\text{{R}} = - d + g - g_c
\end{equation}
\vspace{-0.7cm}
\begin{equation}
\label{eq:mod_rew}
\text{{R$_{SD-DRL}$}} = - d + g - s_c - c_c - c_{c_c} - g_c - c_v - b_{c_c} -ik_c
\end{equation}
where,
\begin{itemize}
    \item $d$: Distance between the end effector and the cube.
    \item $g$: Reward for successfully grasping the object.
    \item $g_c$: Penalty for failed grasp attempt.
    \item $s_c$: Penalty for exceeding a predefined joint speed limit.
    \item $ik_c$: Penalty for inverse kinematic solution failure.
    \item $c_c$: Penalty for collisions with the environment.
    \item $c_{c_c}$: Penalty for collisions with the cube.
    \item $b_{c_c}$: Penalty for collisions with the obstacle.
    \item $c_v$: Penalty for exceeding safety-rated reduced collision velocity ($<250mm/s$ taken from ISO 10218).
\end{itemize}

\subsection{Evaluation Metrics}
The assessment of the SD-DRL utilizes several key evaluation metrics to quantify and compare with the traditional DRL. The metrics evaluated on the agent's behavior are episode returns, safety, and success of the learned policies. 

\subsubsection{Average Episode Returns}
It measures the cumulative sum of the total rewards gained by the agent in one episode. In this study, the average return is divided by the average steps per failed episode to reduce the impact of truncated episodes caused by collisions.

\subsubsection{Average Number of Violations per Episode}
It is used to count the safety compliance of a reinforcement learning agent while performing its operation. It increments the violation counter for events such as collisions, exceeding speed limits, or crossing the velocity threshold for collision safety.

\subsubsection{Success Rate}
It is determined by the agent's ability to achieve its task goals through random states, which in this study is successful grasping that serves as an evaluation for the agent's performance to generalize across different scenarios.

\subsubsection{Safety-Driven Success Rate}
It is a novel metric introduced in this study, which measures an agent's performance to accomplish tasks while complying with safety protocols where a higher score means the agent can achieve goals safely.

\subsection{Safety Standard Compliance}

The study uses the ISO 10218 standards\cite{DINENISO10218-1_2012} and ISO/IEC 61508\cite{DINEN61508-1_2016} regulations which serve as a key part of the methodology. As per ISO 10218, safety-rated speed controls are integrated into the learning objectives in DRL reward behaviors that take safety as a priority like collision avoidance and speed control. On the other hand, the evaluation of the systems is carried out by the ISO/IEC 61508 which is implemented to ensure safety measures in software-controlled systems. The SILs form a fundamental part of implementing the ISO/IEC 61508 guidelines for the measurement of safety functions' effectiveness, which are intended to prevent the risks that exist within the system. This process initiates with conducting risk assessment, which includes utilizing the standard guidelines in conjunction with the metrics assessment to determine the hazards that may be present within the DRL's operational environment. The hazard analysis is the first step of the SIL assignment. This is done by assessing the probability and severity of these hazards and identifying the appropriate SIL for the safety function that matches the level of safety integrity with the magnitude of the risk. This thereby led to the evaluation of functional safety, and the determination of Safety Integrity Levels (SILs) developed SD-DRL algorithm. Validation involves calculating the Probability of Failure on Demand (PFD) and the Risk Reduction Factor (RRF), which are important metrics in SIL determination \cite{DINEN61508-1_2016} given in \cref{eq:PFD}.
\begin{equation} 
\label{eq:PFD}
    PFD \approx (1 - \sum_{s \in S_{mr,lc}} \pi(s)) \times (1 - MTTF)
\end{equation}
\begin{equation}
\label{eq:RRF}
    RRF \approx\frac{1}{PFD}
\end{equation}
where $\pi$ is the approximation of the steady-state probability of safe states, $S_{mr,lc}$ the set of safe operational states, and $MTTF$ the Mean Time To (dangerous) Failure (the total operational steps divided by the number of failures identified as violations). Furthermore, to ensure adherence to safety standards and achieve a robust determination of SILs for our robotic systems' DRL software, we sought guidance from an industrial safety expert in the functional safety domain. 

\section{Design of Experiments}
\label{sec:doe}
The \texttt{(UR5GraspEnv-v0)} focuses on the end effector's position control and not on the direct joint control to ensure the safety of the robot's manipulators. After end effector coordinates have been converted to joint states through inverse kinematics, they are transmitted to the robot controller with safety checks to ensure valid solutions are within the speed limits (soft constraints). If the safety check fails, the controls are ignored.
The workcell collisions or force limits exceeding a 100N threshold are defined as failure (hard constraints - episode termination). Additionally, for failure events during policy testing for velocity during collision validation and functional safety assessment, two changes were made, (i) increasing surface height by 7.5cm and (ii) enlarging object size by 0.5cm. The rewards, determined through assessments using various ranges of values, are shown in \cref{tab:rewards}.
\begin{table}[tbp]
    \caption{Dense rewards and costs for UR5GraspEnv.}
    \begin{tabular}{p{2.9cm}p{0.7cm}p{2.9cm}p{0.7cm}}
    \hline
        \textbf{Parameter} & \textbf{Value} & \textbf{Parameter} & \textbf{Value} \\
    \hline
        \texttt{speed\_cost} & -0.5 & \texttt{\texttt{coll\_vel\_cost}} & -0.5 \\
        \texttt{coll\_cost} & -5.0 & \texttt{gripper\_cost} & -0.01 \\
        \texttt{cube\_coll\_cost} & -0.01 & \texttt{grip\_rew} & 5.0 \\
        \texttt{obstacle\_coll\_cost} & -0.5 & \texttt{grip\_prop\_rew} & 10.0 \\
    \hline
    \end{tabular}
    \label{tab:rewards}
\end{table}
\subsection{Collision (Failure) Avoidance}

\subsubsection{Workspace or Object Collision}
To prevent collisions, the reward function penalizes collisions. If the robot collides with the workspace or the force goes over 100N, the episode terminates.

\subsubsection{Obstacle Collision}
A penalty for obstacle collisions is incorporated into the reward function to encourage the robot to navigate around obstacles while successful task execution. 

\subsection{Speed Reduction at Collision}
A cost is added when velocity exceeds a safety-rated threshold during collisions, aimed to reduce damage and improve safety.
\subsection{Joint Limits}
The specified limits restrict joint movement to ensure safety. Commands are ignored if the speed exceeds safety-standard joint limits (max\_joint\_speed = 2.97 rad for each joint) or for invalid solutions of converting DRL control (end-effector position) to joint positions through inverse kinematics.



\section{Results And Analysis}
\label{sec:results}
\begin{figure}[tbp]
    \centering
        \subfloat[]{
        \includegraphics[width=0.975\linewidth]{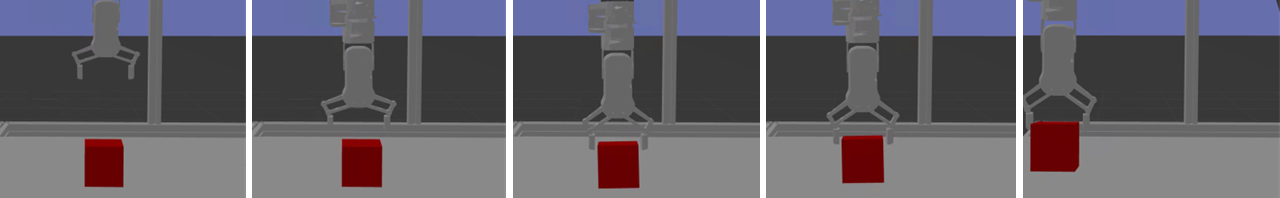}
        \label{}}
            \\
        \subfloat[]{
        \includegraphics[width=0.975\linewidth]{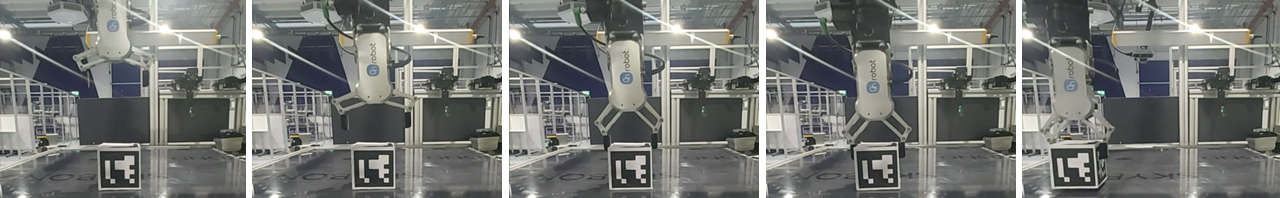}
        \label{}}
            \\
        \subfloat[]{
        \includegraphics[width=0.475\linewidth]{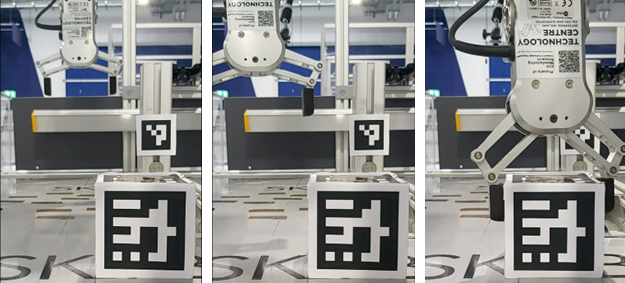}
        \label{}}
        \subfloat[]{
        \includegraphics[width=0.475\linewidth]{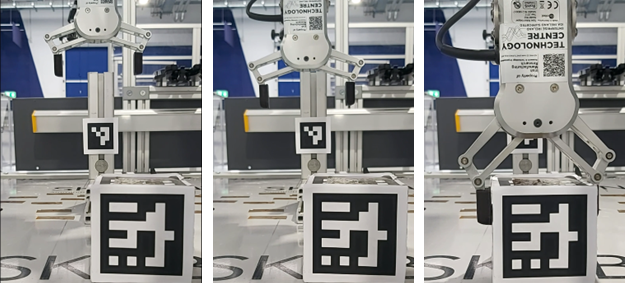}
        \label{}}
            \\
\caption{Policy testing for (a) grasping on simulation, (b) grasping on the testbed, (c) and (d) grasping with obstacle avoidance on the testbed.}
\label{fig:results}
\end{figure}

This section presents the results from simulation training, testing, and real-world validation of the grasping task, comparing the performance of the traditional DRL presented in \cite{katyal2017leveraging} alongside the SD-DRL, with and without obstacles. The policy was trained in simulation for $\approx 2.2 \times 10^6$ steps for the normal scenario (8.3 hours for DRL and 9 hours for SD-DRL) and $\approx 6.5 \times 10^6$ steps for the static obstacle scenario. Trained policy was transferred to the real setup without further training or fine-tuning. During training, 450 tests were conducted for the normal scenario, and 1300 tests for the static obstacle scenario, following every 25 episodes in the simulation. Additionally, 20 tests were executed on the real setup, involving random cube positions for the normal scenario, and random obstacle positions for the obstacle avoidance case. It compares both based on violations, velocity (for simulation), and force (for real setup) during a collision, episode returns, and success metrics. The snapshot for policy validation is shown in \cref{fig:results}. 

\subsection{Violations During Testing in Between Training Steps}
Violations assess the safety performance of the agent. \cref{tab:averages} presents average values for each type of violation during testing. SD-DRL demonstrates fewer violations compared to conventional DRL methods, but in scenarios involving object collision and velocity violation, its incidence tends to be higher. This trend could be attributed to SD-DRL making more successful attempts, resulting in increased interaction with the object. The object collision was merged with the metrics of the collision while validation on the real setup. Furthermore, for additional cross-analysis velocity during the collision in simulation experiments was replaced with force during a collision in the testbed to better understand the hypothesis of indirectly reducing the collision force by reducing the collision velocity. Findings suggest that violations involving physics parameters like speed violation or force during collisions are better for conventional DRL. This implies that fine-tuning the simulation's physics is necessary for developing a safety-driven DRL reward function for real-world applications. As shown in the literature, the impact of the physics engine, ODE may have had an effect on the force during collision and speed violation, which are physics parameters \cite{yoon2023comparative,connolly2021realistic}. Further, the computation of the contact points of the ODE has a significant impact on the nature of the force calculated and the resulting velocity. Therefore, the physics engines, used in the simulation may have an impact on the real-world experiments which is a limitation of Sim2Real for certain scenarios.

\begin{table*}
\centering
\caption{Violations during testing between the training steps and validation on the real setup.}
\label{tab:averages}
\begin{tabular}{p{4cm}p{1.25cm}p{1.25cm}p{1.25cm}p{1.25cm}p{1.25cm}p{1.25cm}p{1.25cm}p{1.25cm}} 
\hline
\multirow{3}{*}{\textbf{Violation Type}} & \multicolumn{4}{c}{\textbf{Normal Scenario}}                                                                                                                                            & \multicolumn{4}{c}{\textbf{Static Bar Obstacle Scenario}}                                                                                                                                                                        \\ 
\cline{2-9}
                                         & \multicolumn{2}{c}{\textbf{Simulation }}           & \multicolumn{2}{c}{\textbf{Real-World}}                                                                                    & \multicolumn{2}{c}{\textbf{\textbf{Simulation}}}                                       & \multicolumn{2}{c}{\textbf{\textbf{Real-World}}}                                                                            \\ 
\cline{2-9}
                                         & \textbf{\textbf{DRL}} & \textbf{\textbf{SD-DRL}} & \textbf{\textbf{\textbf{\textbf{DRL}}}} & \textbf{\textbf{\textbf{\textbf{\textbf{\textbf{\textbf{\textbf{SD-DRL}}}}}}}} & \textbf{\textbf{\textbf{\textbf{DRL}}}} & \textbf{\textbf{\textbf{\textbf{SD-DRL}}}} & \textbf{\textbf{\textbf{\textbf{\textbf{\textbf{\textbf{\textbf{DRL}}}}}}}} & \textbf{\textbf{\textbf{\textbf{SD-DRL}}}}  \\ 
\hline
Collision                                & 0.048                 & \textbf{0.043}             & \textbf{0.200}                          & 0.400                                                                            & 1.548                                   & \textbf{0.162}                               & 0.737                                                                       & \textbf{0.421}                                \\
Obstacle Collision                       & \textbf{0.002}        & 0.014                      & -                                       & -                                                                                & 17.54                                   & \textbf{14.24}                               & 0.316                                                                       & \textbf{0.053}                                \\
Speed Violation                          & \textbf{0.018}        & \textbf{0.018}             & \textbf{0.200}                          & 1.050                                                                            & 0.163                                   & \textbf{0.158}                               & \textbf{0.263}                                                              & 2.053                                         \\
Velocity Violation                       & \textbf{0.037}        & 0.043                      & -                                       & -                                                                                & 14.95                                   & \textbf{5.511}                               & -                                                                           & -                                             \\
Velocity During Collision                & 0.439                 & \textbf{0.231}             & -                                       & -                                                                                & 0.264                                   & \textbf{0.209}                               & -                                                                           & -                                             \\
Force During Collision                   & -                     & -                          & \textbf{34.89}                          & 37.11                                                                            & -                                       & -                                            & \textbf{24.12}                                                              & 31.98                                         \\
\hline
\end{tabular}
\end{table*}

\subsection{Velocity Profiles During Collision}
\begin{figure}[bp]
    \centering
        \subfloat[]{
        \includegraphics[width=0.48\linewidth]{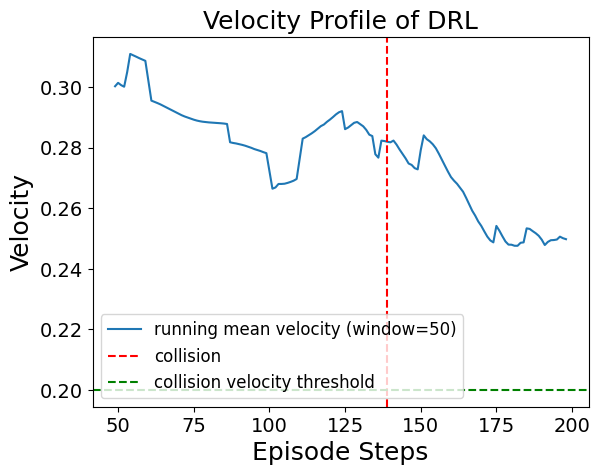}
        \label{fig:vel_pro_drl_1}}
        \subfloat[]{
        \includegraphics[width=0.48\linewidth]{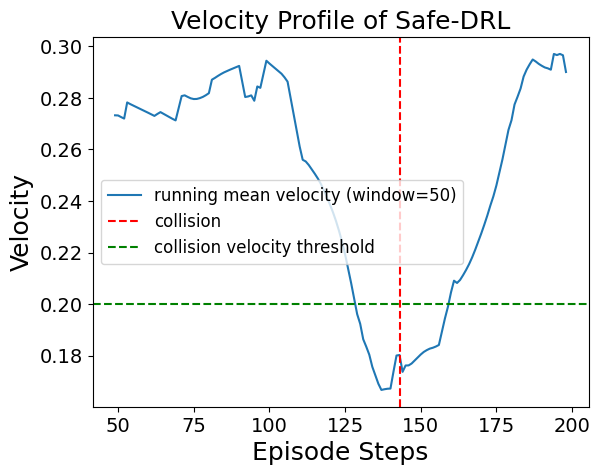}
        \label{fig:vel_pro_safe_drl_1}}
        \label{fig:vel_profiles}
\caption{Velocity profiles for (a) DRL and (b) Safety-Driven DRL during a collision.}
\label{fig:vel_profiles}
\end{figure}

Analysis of velocity profiles (\cref{fig:vel_profiles}) during collisions validates SD-DRL's safe behavior compared to conventional DRL. In conventional DRL (\cref{fig:vel_pro_drl_1}), the robot arm's velocity shows abrupt or no change upon collision, indicating a lack of collision anticipation, potentially leading to damage. In contrast, SD-DRL (\cref{fig:vel_pro_safe_drl_1}) demonstrates smoother velocity transitions, suggesting collision anticipation and adjustment to minimize impact force. This highlights SD-DRL's ability to balance task completion and safety, crucial for reliable and responsible operation in real-world environments, especially in physically interactive applications.

\subsection{Success Rate and Safety-Driven Success Rate}

The success rate measures how often a reinforcement learning agent achieves its task goals. Safety-driven success rate adds adherence to safety constraints to this metric. SD-DRL outperforms conventional DRL based on these metrics, as shown in \cref{tab:safety_success_rate}.

\begin{table*}
\centering
\caption{Success during testing between the training steps and validation on the real setup.}
\label{tab:safety_success_rate}
\begin{tabular}{p{4cm}p{1.25cm}p{1.25cm}p{1.25cm}p{1.25cm}p{1.25cm}p{1.25cm}p{1.25cm}p{1.25cm}} 
\hline
\multirow{3}{*}{\textbf{Success Metrics}} & \multicolumn{4}{c}{\textbf{Normal Scenario}}                                                                                                                                            & \multicolumn{4}{c}{\textbf{Static Bar Obstacle Scenario}}                                                                                                                                                                        \\ 
\cline{2-9}
                                          & \multicolumn{2}{c}{\textbf{Simulation }}           & \multicolumn{2}{c}{\textbf{Real-World}}                                                                                    & \multicolumn{2}{c}{\textbf{\textbf{Simulation}}}                                       & \multicolumn{2}{c}{\textbf{\textbf{Real-World}}}                                                                            \\ 
\cline{2-9}
                                          & \textbf{\textbf{DRL}} & \textbf{\textbf{SD-DRL}} & \textbf{\textbf{\textbf{\textbf{DRL}}}} & \textbf{\textbf{\textbf{\textbf{\textbf{\textbf{\textbf{\textbf{SD-DRL}}}}}}}} & \textbf{\textbf{\textbf{\textbf{DRL}}}} & \textbf{\textbf{\textbf{\textbf{SD-DRL}}}} & \textbf{\textbf{\textbf{\textbf{\textbf{\textbf{\textbf{\textbf{DRL}}}}}}}} & \textbf{\textbf{\textbf{\textbf{SD-DRL}}}}  \\ 
\hline
Success rate                              & 0.29                  & \textbf{0.36}              & 0.15                                    & \textbf{0.25}                                                                    & 0.00                                    & \textbf{0.30}                                & 0.00                                                                        & \textbf{0.16}                                 \\
Success attempts                       & 125                   & \textbf{157}               & 3                                       & \textbf{5}                                                                       & 0                                       & \textbf{405}                                 & 0                                                                           & \textbf{3}                                    \\
Safety-driven success rate             & 0.28                  & \textbf{0.34}              & 0.15                                    & \textbf{0.15}                                                                    & 0.00                                    & \textbf{0.27}                                & 0.00                                                                        & \textbf{0.05}                                 \\
Safety-driven success attempts      & 121                   & \textbf{150}               & \textbf{3}                              & \textbf{3}                                                                       & 0                                       & \textbf{375}                                 & 0                                                                           & \textbf{1}                                    \\
Average return                            & 0.53                  & \textbf{0.89}              & -0.23                                   & \textbf{-0.21}                                                                   & -0.37                                   & \textbf{-0.07}                               & -0.38                                                                       & \textbf{-0.33}                                \\
\hline
\end{tabular}
\end{table*}

\section{Functional Safety Assessment}
\label{sec:fsa}
The Functional Safety Assessment (FSA) involves a risk assessment and SIL determination which calculates the MTTF, PFD, and RRF. These metrics were derived from operational data gathered during both simulated and real-world implementation of the DRL and SD-DRL system for both the cases of normal and static obstacle scenarios combined. Results were extracted from the simulation with induced disturbance (discussed in \cref{sec:doe}) using the trained policy for 500 episodes and the same real-world data was used for which the results are reported in \cref{sec:results}. The MTTF was calculated as the total operational steps divided by the number of failures identified as collision and speed violations. The PFD was estimated based on the frequency of demand for safety-critical functions and the RRF was determined as the inverse of the PFD. \cref{tab:sil_values} presents a comparative analysis of SIL determination across different setups. The final SIL determination was made considering PFD and risk assessment carried out under guidelines by industrial safety experts, as it ensures that the SIL determination is reflective of both empirical data and expert evaluation of the system’s operational safety. For our specific scenario, we assume pi is equal to 0, as the current operational characteristics ensure failures are non-hazardous. Both models achieve an SIL 2, as determined by a safety expert and based on the PFD.  However, the improved metrics in the safety-driven model highlight the effectiveness of additional safety measures implemented in this setup. The DRL-based system demonstrates substantial compliance with required safety standards, achieving a consistent Safety Integrity Level of 2 across different testing environments. These results underscore the potential of DRL systems to enhance functional safety in complex and interactive manufacturing settings. 

\begin{table}
\centering
\caption{SILs from the combined experimental results involving normal and static obstacle scenario.}
\label{tab:sil_values}
\begin{tabular}{p{1cm} p{0.6cm} p{1.25cm} p{0.6cm} p{1.25cm} p{1.5cm}}
\hline
\multirow{2}{*}{\textbf{Metrics}} & \multicolumn{2}{l}{\textbf{\textbf{Simulation}}} & \multicolumn{2}{l}{\textbf{\textbf{Real-World}}} & \multirow{2}{*}{\textbf{SIL 2 Range}}  \\ 
\cline{2-5}
                                  & \textbf{DRL}    & \textbf{\textbf{SD-DRL}}     & \textbf{DRL} & \textbf{SD-DRL}                 &                                        \\ 
\hline
MTTF                              & \textbf{593.53} & 549.85                         & 440.30       & \textbf{742.34}                   & \textgreater{100} steps                              \\
PFD                               & \textbf{0.0017} & 0.0018                         & 0.0023       & \textbf{0.0013}                   & 0.01 to 0.001  \\
RRF                               & \textbf{593.53} & 549.85                         & 440.30       & \textbf{742.34}                  & 100 to 1000                            \\
\hline
\end{tabular}
\end{table}

\section{Conclusions and Future Work}
\label{sec:conc}
This study demonstrated the successful integration of safety compliance into the reward function of the DRL algorithm and proposed a new framework called Safety-Drived DRL. This framework identifies and avoids potential hazards across two main scenarios i.e. grasping objects with and without obstacles. Through simulation, a model was trained, validated, and deployed on a real-world setup, where the algorithm was tested for its operational efficiency. Further, its functional safety was validated across all the scenarios using the IEC 61508. This assessment showed the improvements of the proposed SD-DRL over traditional DRL while maintaining operational efficiency. While some of the SD-DRL results related to the physics-based parameters did not meet the real-world results which were due to the choice of the physics engine in Gazebo, the study aims to extend further with testing with various other physics engines such as Bullet. Also, future studies will involve the use of physics-informed neural networks for improvements in the performance of the SD-DRL in safety-critical robotic systems for Sim2Real approaches. The validation process included determining Safety Integrity Levels (SILs) for DRL systems, essential for ensuring compliance with safety standards in safety-critical environments. However, maintaining consistent safety standards posed challenges due to the adaptive nature of DRL, necessitating periodic evaluations of safety performance. Nonetheless, the study concluded that SD-DRL not only optimized tasks but also significantly enhanced functional safety in robotics, emphasizing the importance of fine-tuning simulator parameters to match real-world conditions for future research. Future work involves prediction and avoidance of violations and includes a case study related to dynamic human collision avoidance, advancing safe human-robot collaboration. 

\section*{Acknowledgements}
This work is supported by the EU-funded CISC project (MSCA grant: 955901), partly funded by ADAPT Research Center, Ireland, and the State of Upper Austria through the SCCH INTEGRATE (FFG grant: 892418). Special thanks to the robotics team (Sunny Katyara, Ted Morell, Declan O'Neill, Keerthi Sagar, Aayush Jain, Francis O'Farrell, and Court Edmondson) at Irish Manufacturing Research and Islam Attia from Pilz Industrial Automation, Ireland for their contributions.


\bibliographystyle{IEEEtran}
\bibliography{bibliography}

\end{document}